%% file: main.tex
\documentclass[10pt,twocolumn,letterpaper]{article}

\usepackage[pagenumbers]{cvpr} 


\usepackage{latexsym}
\usepackage{amsmath,amssymb,amsfonts}
\usepackage{amsthm}
\usepackage{bbm}
\usepackage{booktabs}
\usepackage{enumitem}
\usepackage{graphicx}
\usepackage{color}
\usepackage{url}
\usepackage{algorithm}
\usepackage{arydshln}
\usepackage{algpseudocode}
\usepackage{subcaption}
\usepackage{breqn}
\usepackage{comment}
\usepackage{makecell}

\makeatletter
\def\adl@drawiv#1#2#3{%
        \hskip.5\tabcolsep
        \xleaders#3{#2.5\@tempdimb #1{1}#2.5\@tempdimb}%
                #2\z@ plus1fil minus1fil\relax
        \hskip.5\tabcolsep}
\newcommand{\cdashlinelr}[1]{%
  \noalign{\vskip 0.2\aboverulesep
           \global\let\@dashdrawstore\adl@draw
           \global\let\adl@draw\adl@drawiv}
  \cdashline{#1}
  \noalign{\global\let\adl@draw\@dashdrawstore
           \vskip 0.2\belowrulesep}}
\makeatother

\definecolor{cvprblue}{rgb}{0.21,0.49,0.74}
\usepackage[pagebackref,breaklinks,colorlinks,allcolors=cvprblue]{hyperref}
\def\paperID{*****} 
\def\confName{CVPR}
\def\confYear{2025}

\title{ERDE: Entropy-Regularized Distillation for Early-exit}

\author{Martial Guidez\\
\small INSA Lyon, CNRS, Université Claude Bernard Lyon 1,\\
\small LIRIS, UMR5205, 69621 Villeurbanne, France\\
{\tt\small martial.guidez@liris.cnrs.fr}
\and
Stefan Duffner \\
\small INSA Lyon, CNRS, Université Claude Bernard Lyon 1,\\
\small LIRIS, UMR5205, 69621 Villeurbanne, France\\
{\tt\small stefan.duffner@liris.cnrs.fr}
\and
Yannick Alpou \\
{\tt\small yannickalpou@gmail.com}\\
\small INSA Lyon, CNRS, Université Claude Bernard Lyon 1,\\
\small LIRIS, UMR5205, 69621 Villeurbanne, France\\
\and
Oscar Röth \\
{\tt\small oscar.roeth@gmail.com}\\
\small INSA Lyon, CNRS, Université Claude Bernard Lyon 1,\\
\small LIRIS, UMR5205, 69621 Villeurbanne, France\\
\and
Christophe Garcia \\
\small INSA Lyon, CNRS, Université Claude Bernard Lyon 1,\\
\small LIRIS, UMR5205, 69621 Villeurbanne, France\\
{\tt\small 	christophe.garcia@liris.cnrs.fr}
}

\begin{document}
\maketitle

\input{0_abstract}    
\input{1_intro}
\input{2_relatedwork}

\input{3_method}
\input{4_expe}
\input{5_results}
\input{7_conclusion}
\input{8_acknowledgements}

{
    \small
    \bibliographystyle{ieeenat_fullname}
    \bibliography{main}
}


\input{9_appendix}

\end{document}

%% file: 0_abstract.tex
\begin{abstract}
Although deep neural networks and in particular Convolutional Neural Networks have demonstrated state-of-the-art performance in image classification with relatively high efficiency,
they still exhibit high computational costs, often rendering them impractical for real-time and edge applications. 
Therefore, a multitude of compression techniques have been developed to reduce these costs while maintaining accuracy.
In addition, dynamic architectures have been introduced to modulate the level of compression at execution time, which is a desirable property in many resource-limited application scenarios.
The proposed method effectively integrates two well-established optimization techniques: early exits and knowledge distillation, where a reduced student early-exit model is trained from a more complex teacher early-exit model. 
The primary contribution of this research lies in the approach for training the student early-exit model.
In comparison to the conventional Knowledge Distillation loss, our approach incorporates a new entropy-based loss for images where the teacher's classification was incorrect.
The proposed method optimizes the trade-off between accuracy and efficiency, thereby achieving significant reductions in computational complexity without compromising classification performance. 
The validity of this approach is substantiated by experimental results on image classification datasets CIFAR10, CIFAR100 and SVHN, which further opens new research perspectives for Knowledge Distillation in other contexts.
\end{abstract}

%% file: 1_intro.tex
\section{Introduction}

In the field of deep learning, the reduction of computational cost is a matter of significant concern. A multitude of techniques have been identified that adapt Convolutional Neural Network (CNN) architectures, thereby reducing both computational cost and model size. These techniques have proven to be efficient in various aspects of architecture optimization, where the combination of methods is a particularly effective approach to achieving high compression rates when they are complementary, such as a pruning and a quantization method, for example.

For instance, Qi  \etal \cite{Qi2021LearningLR} achieve a compression rate of 50\% with an accuracy reduction of only 0.15\%-0.37\%. Another method proposed a combination early-exit and quantization on CNN \cite{Li2022PredictiveEP} reaching a compression reduction of 50\% with an accuracy drop of 1\% to 3\%. 
Quantization can also be effectively combined with pruning, e.g.~\cite{Han2016DeepCompression}.
Thus, the combination of distinct methods allows to further optimize the trade-off between compression and accuracy. 

In this study, we integrate the knowledge distillation and early-exit approaches to obtain distilled dynamic neural networks. That is, at inference time, the distilled network can further reduce its compression rate dynamically by executing only parts of the model depending on external conditions (e.g. battery power) or internal criteria (e.g.\ the difficulty of the input).

Knowledge distillation is a training method in which a large reference network, designated as the "teacher", is employed to train a smaller network, referred to as the "student". 
This approach enables the student network to attain a higher accuracy compared to training without KD. 
Early-exit (or multi-exit) methods employ a dynamic compression technique where the network makes multiple intermediate predictions after executing a certain number of layers. 
The complexity of the model can thus be modulated (automatically or manually) at run-time by exiting earlier from the neural network and thereby eliminating the need for subsequent calculations.

The proposed method named ERDE for Entropy-Regularized Distillation for Early-exit relies on a specific training method that allows us to apply knowledge distillation between two early-exit networks. 
By effectively combining the two approaches, we are able to cumulate the compression gains of both and, in addition, obtain a run-time configurable model.

In summary, our contributions are the following:
\begin{itemize}
\item We present a new compression approach that effectively combines Knowledge Distillation with Early-Exit dynamic neural network architectures.
\item We introduce a new distillation method based on a loss function applied at intermediate exits trying to maximize the entropy for those examples where the teacher model is uncertain.
\end{itemize}

We have tested our approach with different CNNs on three classical image classification benchmarks.
Our models obtained a significant reduction in computational complexity, i.e.\ up to around 10 times lighter than the original models, with little loss in accuracy or even a gain in some cases. 
In addition, compared to the standard KD training algorithm, applying KD to EE architectures with our proposed entropy-regularization loss improves the average accuracy on all tested datasets for all possible EE thresholds.

%% file: 2_relatedwork.tex
\section{Related Work}

\subsection{Dynamic Neural Networks}

The need for dynamic networks comes from the inability of static ones to adapt the computational graphs or the network parameters. For example, processing complex images requires deep and complex networks that require more computation. However, this type of network will do a large amount of unnecessary computation when processing simpler images.

Many techniques have been developed to adapt the architecture \cite{Han2021DynamicNN} of neural networks at execution time. We can divide these methods into three categories: dynamic depth, dynamic width, and dynamic routing. Each category in turn contains several techniques.

Dynamic depth methods are designed to avoid the redundant computations of deep networks mentioned above. This can be achieved by exiting at shallow exits for simple inputs (early exit)~\cite{RahmathP2024EarlyExitDN} or by selectively skipping certain intermediate layers given a particular input (layer skipping)~\cite{Graves2016AdaptiveCT,Wang2017SkipNetLD}.
Dynamic width methods are more fine-grained than the previous ones, all layers are executed but some units (neurons, channels or branches) are not activated depending on the input. One of the dynamic width methods is Mixture of Experts (MoE)~\cite{Jacobs1991AdaptiveMO,Eigen2013LearningFR}. It is a strategy that dynamically selects and uses only a small part of the network (the experts) for processing, based on the given input.
Dynamic routing is a key mechanism in modern dynamic neural networks that adapt their computational pathways based on input. It enables the network to optimize its structure by selectively activating specific layers, channels, or paths within architectures like SuperNets \cite{Cha2022SuperNetIN}
. We can consider early-exiting networks as a special form of SuperNets. Another type of dynamic routing networks are CapsuleNets, where routing between capsules captures hierarchical relationships~\cite{sabour2017dynamic,hinton2018matrix}.

\subsection{Early-Exit Approaches}

Early-exit methods add various exits at different layers and are composed of additional classification branches (denoted ``exit branches'' in the following). 
The inference is initially performed up to the first exit branch. At this point, a prediction is made with an associated confidence. If the confidence reaches a certain threshold, the inference is stopped and the resulting prediction is returned, otherwise the inference continues until the next exit.

The exit branches can be composed of one or more different layers: 
either a classic single Fully Connected (FC) layer~\cite{Chen2020LearningTS,Dai2020EPNetLT}, multiple FC layers~\cite{Zhao2021EdgeMLAA} or more complex architectures 
adding one or more convolution layers and pooling layers before the FC layer~\cite{Teerapittayanon2016BranchyNetFI, Jo2023LoCoExNetLE}.
Note that the calculations performed on these exit branches are "lost" if the confidence is below the threshold and thus the exit not used, so it is important to ensure that the exits remain cost effective. 
The placement of the exit branches may also differ. 
Some methods add an exit after each layer \cite{Dai2020EPNetLT}, others choose specific layers \cite{Berestizshevsky2018DynamicallySA,Teerapittayanon2016BranchyNetFI} and finally there are more complex techniques like using metrics or gating function to decide where to put the exits~\cite{Fang2020FlexDNNIO,Li2023EENetEE}.

The inclusion of additional exits to a classical DNN entails a different training strategy because the network has several outputs that need to be optimized. 
Joint training is the most trivial method, in which all branches are trained simultaneously~\cite{Teerapittayanon2016BranchyNetFI}, where the global loss is defined as the sum of all losses obtained at the end of each branch considered by a chosen coefficient. In the branch-wise training strategy, each side branch is trained separately, together with the preceding layers of the backbone DNN \cite{belilovsky2019shallow}. 
Separate training consists of treating the exit branches as independent classifiers and training them independently~\cite{Chiu2023FairMF}. 
Another training strategy, called two-stage training, is to train the DNN backbone first, then its parameters are frozen and the exit branches are trained separately~\cite{Berestizshevsky2018DynamicallySA}. 
Knowledge Distillation (KD) based training uses the exit branches as student models that learn from the output of the DNN backbone~\cite{He2015DelvingDI,Hinton2014DistillingTK}.

Furthermore, many exit policies have been developed to decide whether to exit or not depending on the branch output. These policies can be divided into two categories: static (rule-based) policies and dynamic (learnable) policies.
Static exit policies measure the confidence in the predictions made using metrics such as entropy, maximum softmax or user-defined scoring functions. A single threshold can be applied to all branches, or the threshold can be specified for each branch. These methods are easy to implement and fast, but lack robustness due to their inability to adapt. 
On the other hand, some learnable exit policies have been proposed, such as exit selection controllers~\cite{Dai2020EPNetLT,Guan2017EnergyefficientAI}, reinforcement learning~\cite{Guan2017EnergyefficientAI,Wang2017SkipNetLD}, and soft gating mechanisms~\cite{Mullapudi2018HydraNetsSD}. 
They have a high training complexity but also a high robustness to diverse input. 

\subsection{Distillation and Dynamic Neural Networks}

One of the commonly used compression methods is called Knowledge Distillation (KD)~\cite{Hinton2014DistillingTK}. It consists in using a reference model (teacher) to train a smaller network (student), leading to a better accuracy in a shorter time compared to the same student trained without KD. 
%
Inspired by KD, L. Zhang \etal \cite{Zhang2019BeYO} have developed self-distillation, which unlike traditional KD, works within a single network. It is a training technique to improve model performance, not a method to compress models. The idea of this method is to divide the network into different sections and distill the knowledge from the deeper sections to the shallower ones. This approach not only improves accuracy but also promotes computational efficiency.

Building on the principles of self-distillation, other researchers have explored its potential in terms of optimisation. For example, ESCEPE~\cite{KhalilianGourtani2023ESCEPEEN} is an approach based on weight clustering and self-distillation achieving a high compression ratio of the early-exit network with minimal impact on the accuracy of the intermediate classifiers.
The proposed method combines early-exit with self-distillation: a special case of knowledge distillation where the teacher is the network before compression. As apposed to our approach, this work does not explore the possibility of using a different network as a teacher and focuses mainly on pruning a network.

%% file: 3_method.tex
\section{Method}

\begin{figure*}
    \centering
    \includegraphics[width=1\linewidth]{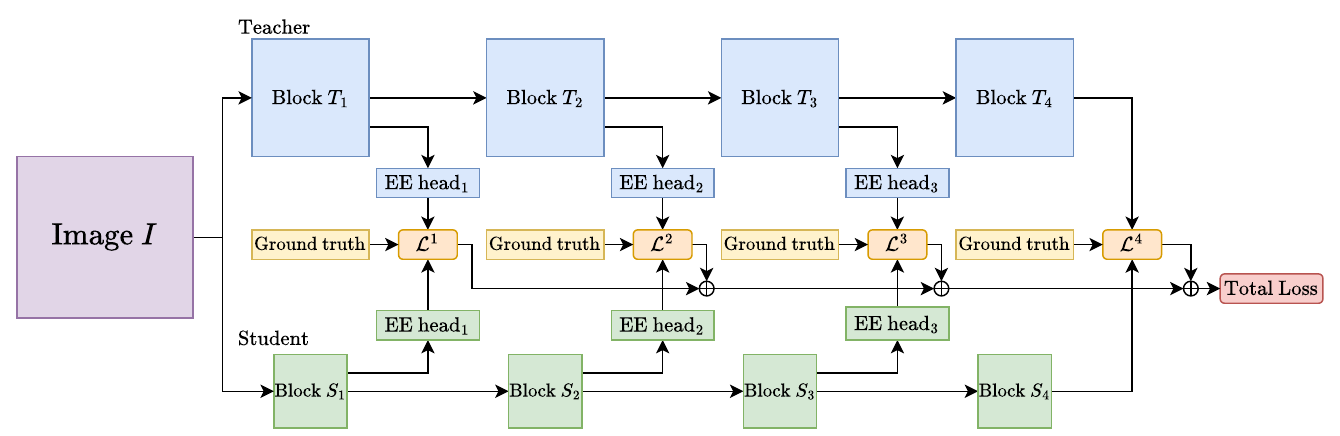}
    \caption{Our ERDE architecture and training approach for 4 teacher and student blocks (in blue and green respectively). The $\mathcal{L}^i$ correspond to the different loss functions ($\mathcal{L}_{\text{CE}}$, $\mathcal{L}_{\text{KD}}$, and $\mathcal{L}_{\text{E}}$) at the $i$-th exit. 
    At inference, only the compressed student model is used.}
    \label{fig:ERDE}
    \vspace{-1em}
\end{figure*}

This section delineates the methodology utilized and the contribution of the present study. 
The proposed architecture (illustrated in \autoref{fig:ERDE}) uses early exits with $n-1$ branches, i.e.\ there are $n-1$ early exit heads, and the $n$-th one is the classical exit of the network. 
The architecture of the EE heads will be defined in the subsequent section. As previously outlined, the premise of early exits is that the example progresses through the network until the first exit. At this point, a primary prediction is conducted, accompanied by the attribution of an uncertainty score $c_i$ to the prediction. This score is defined as the entropy of the softmax vector of the prediction.
If the uncertainty score is below  a predetermined threshold, the prediction is designated as the network's output, i.e., the final prediction. This predetermined threshold, denoted as $\theta$ in $[0;\infty[$ , will be consistent for all exits.
As the threshold diminishes, a greater proportion of the samples will depart from the lowest exit. 
Conversely, as the threshold increases, a higher percentage of the samples will exit from the deeper exits.
If the score falls below the threshold, the image $I$ will persistently traverse the network until the subsequent exit, a process that is repeated. 
If the network does not hold sufficient confidence in a prediction at the $n-1$-th exit, the image will terminate at the ultimate exit (corresponding to the conventional end of the network). 
In this study, entropy was selected as the metric of confidence, a common choice in the literature  \cite{Teerapittayanon2016BranchyNetFI}). 

We did not explore further EE architectures or policies because we believe this is not part of our contribution. Algorithm \ref{alg:early-exit} formalizes the early-exit inference mechanism.

\begin{algorithm}[ht]
    \begin{algorithmic}
    \State $i \gets 1$
    \State $h_0 \gets I$ 
    \While{$c_i > \theta$ and $i \leq n$}
    \State $h_{i} \gets \text{Block}_i(h_{i-1}$)
    \State $y_i \gets \text{EE head}_i(h_i)$
    \State $c_i \gets$ entropy(softmax($y_i$))
    \State $i ++$
    \EndWhile
    \State \Return $y_i$
    \end{algorithmic}
 \caption{Early-exit inference algorithm.}
 \label{alg:early-exit}
\end{algorithm}

The training of a multi-exit network necessitates a modification of the conventional training process. This modification is required to ensure the training of all heads and the calculation of the confidence threshold, which is a prerequisite in certain cases. The objective is to consider the various exit losses. The subsequent paragraph provides a detailed definition of these losses.

Firstly, the following notation must be established: $y_{T_i} = sigmoid(z_{T_i},T)$, the softmax of the $i$-th exit of the teacher; $y_ {S_i} = sigmoid(z_ {S_i},T)$, the softmax of the $i$-th exit of the student. Where $T$ is the temperature parameter of the softmax function. Assume, that the number of classes of our problem is $K$ and the number of branches of both, teacher and student, is $n$.

The classic loss employed in KD is defined as the sum of two losses: the KD loss and the CE loss. The KD loss is expressed as the difference between the student's output and the teacher's output. This loss is designed to encourage the student to make a prediction that mirrors the teacher's prediction. It can be defined through various functions; however, the Kullback-Leibler divergence, which was utilized in the original work by G. Hinton \cite{Hinton2014DistillingTK}, was selected for implementation in this study: 
\begin{equation}
    \mathcal{L}^i_{\text{KL}} 
    = \frac{T^2}{K} \times \sum_{k\in K} \hat{y}_{T_{i_k}} 
    \log\left(\frac{\hat{y}_{T_{i_k}}}{\hat{y}_{S_{i_k}}} \right) \; .
\label{eq:lossKL}
\end{equation}

The Cross-Entropy (CE) loss is defined as the cross entropy between the ground truth and the student prediction

\begin{equation}
        \mathcal{L}^i_{\text{CE}} = - \sum_{k\in K} y_k \log\left(\hat{y}_{S_{i_k}}\right)  \; .
\label{eq:lossCE}
\end{equation}

This loss compared the class given by the student with the ground truth. $\mathcal{L}^i_{\text{CE}}$ is also used to train the teacher network in the following experiments.\medskip

When thinking of combining knowledge distillation and early exit, our idea was to adapt the loss to take in account the cases where the teacher failed to accurately classified the input. For the samples that are correctly classified by the teacher, we use the classic 

In order to train the student and take into account the early exit, we adapted the classic knowledge distillation method. 
The primary contribution of the proposed method is the division of the loss between two cases: one for samples that are correctly classified by the teacher and one for samples that are incorrectly classified. 
In the event that the teacher has successfully classified the sample, we use the combination of $\mathcal{L}_{\text{KL}}$ and $\mathcal{L}_{\text{CE}}$ as usual. In the alternative scenario, the entropy of the output vector is subtracted. Specifically, this entails the subtraction of a quantity $\mathcal{L}_{\text{E}}$ represented as:

\begin{equation}
        \mathcal{L}^i_{\text{E}} = \sum_{k\in K}\hat{y}_{S_{i_k}} \log\left(\hat{y}_{S_{i_k}}\right)  \; .
\label{eq:lossE}
\end{equation}

The idea of this entropy loss is to force the network to be "uncertain" for samples where the teacher failed to accurately predict the class of the image. 
The maximum of this loss for a sample is achieved when all classes are equally predicted and is minimized when a class is predicted with a probability of 1. 

Finally the overall loss can be written as:
\begin{equation}
    \begin{split}
        \mathcal{L}_{tot} = & \sum^{n-1}_{i=0}\mathbbm{1}_{y=\hat{y}_{T_{i}}}(\omega_{\text{KL}}  \mathcal{L}^i_{\text{KL}} + \omega_{\text{CE}}  \mathcal{L}^i_{\text{CE}}) - \mathbbm{1}_{y\neq\hat{y}_{T_{i}}} \omega_{\text{E}} \mathcal{L}^i_{\text{E}} \\
        & + \omega_{\text{KL}}  \mathcal{L}^n_{\text{KL}} + \omega_{\text{CE}}  \mathcal{L}^n_{\text{CE}} \; .
    \label{eq:lossTOT}    
    \end{split}
\end{equation}

At the final exit, we employ the conventional KD loss regardless of the accuracy of the teacher prediction. 
This choice stems from the principle that the final exit is required to produce an output, even for instances where the confidence threshold is not attained. 
In contrast, the purpose of our entropy loss is to accentuate the uncertainty in intermediate exits, thereby preventing erroneous intermediate predictions.

%% file: 4_expe.tex
\section{Experiments}

We evaluated our approach on three different standard datasets for image classification: CIFAR10, CIFAR100\footnote{https://www.cs.toronto.edu/~kriz/cifar.html} and SVHN\footnote{http://ufldl.stanford.edu/housenumbers}%
\footnote{Ror SVHN, we did not use the optional extra training data in our experiments.} and
 on a different student-teacher couples including ResNet~\cite{he2016resnet} and ConvNeXT architectures~\cite{Liu2022ACF}. 
We have used 3 different ResNet networks 34, 10, and 8. Where ResNet8 is similar to a ResNet10 but with only 3 blocks (instead of 4 in a ResNet10). 
We added an exit branch after each block, where the last exit corresponds to the final output of the full model. 
To minimise the computational overhead, our exit branches are very shallow and operate on the last convolution layer of the preceding block. 
They contain only a batch normalization layer, a ReLU activation function, a 2x2 average pooling layer, a Dropout with probability 0.5 and a single FC layer that performs the intermediate and final predictions. 

For each of the datasets, we compared our model trained with our proposed distillation loss (\ref{eq:lossTOT}) to the following baselines: a teacher model without early exits, a student without KD and a student trained with classical KD using the sum of the cross-entropy loss and the distillation loss (KL divergence). 

We trained the early-exit models by simultaneously minimizing the loss for all exits, i.e.\ we simply minimized the sum of all losses without any weighting. 
The teacher and student models without KD are trained with the CE loss (\ref{eq:lossCE}) without any (self-)distillation.
The student models with classical KD are trained with
\begin{equation}
    \mathcal{L}_{\text{KD}} = \sum^{n}_{i=0}\omega_{\text{KL}}  \mathcal{L}^i_{\text{KL}} + \omega_{\text{CE}}  \mathcal{L}^i_{\text{CE}}  \; ,
\label{eq:lossStudentNoKD}    
\end{equation}
and to evaluate our proposed approach we trained models with the loss in (\ref{eq:lossTOT}).

All models are trained for 300 epochs with a batch size of 64 and a learning rate of $10^{-3}$ with the Adam algorithm.
To avoid overfitting, we applied early stopping with a validation set of 5000 images for CIFAR10 and CIFAR100 and 13256 images for SVHN.
For the ConvNeXT models we started with ImageNet pre-trained weights, otherwise the models were not able to converge properly and did overfit too much.
Furthermore, data augmentation is performed with random horizontal flips, rotations, translations, crops and random erasing.
For all experiments, we set $\omega_{\text{KL}}=0.25$, $\omega_{\text{CE}}=0.75$, $\omega_E=0.005$ and $T=2$. 

For evaluating the different models, we computed the accuracy and the number of MACs (Multiply-Accumulate operations) and varied the confidence thresholds for the early-exit models.
As the number of operations varies for each example, we report the average MACs (per example) over the whole test set.
The average latencies are computed with a batch size of 1 on a NVIDIA V100 graphics card including all data loading and transfer overhead. 

A part from standard KD, we did not compare to other network compression methods, as they are not directly comparable but they are rather complementary. For example, in pruning approaches, the compression ratio can usually be chosen in a fine-grained manner, whereas in KD we fix a target student architecture in advance.

%% file: 5_results.tex
\section{Results}

The overall performance results in terms of accuracy, MMACs and latency is shown in \autoref{tab:cifar10}, \ref{tab:cifar100}, and \ref{tab:svhn}.
The teacher accuracy is usually higher than the one of the student model trained without KD.

\begin{figure}[ht]
    \centering
    \begin{subfigure}{\linewidth}
        \includegraphics[width=1\textwidth]{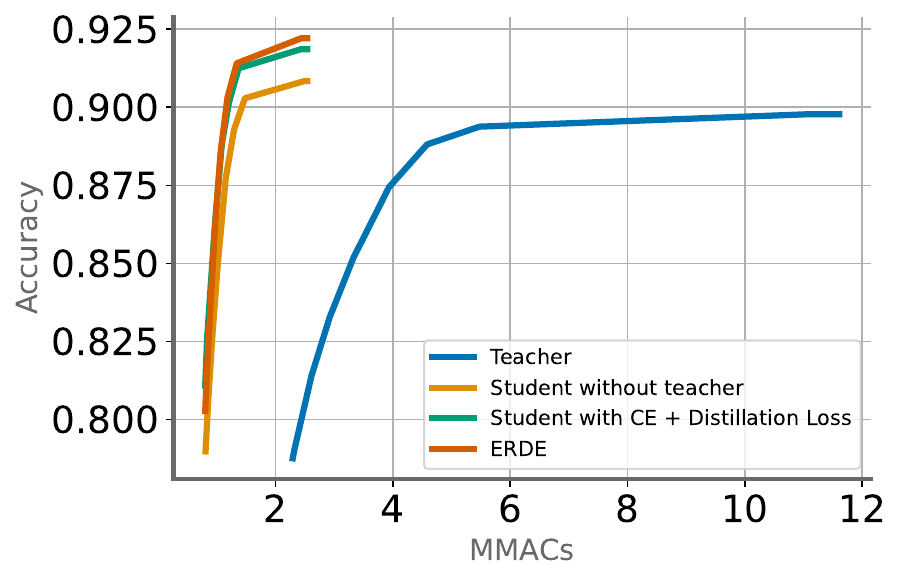}
        \caption{CIFAR10}
        \label{fig:cifar10}
    \end{subfigure}\hfill
    \begin{subfigure}{\linewidth}
        \includegraphics[width=1\textwidth]{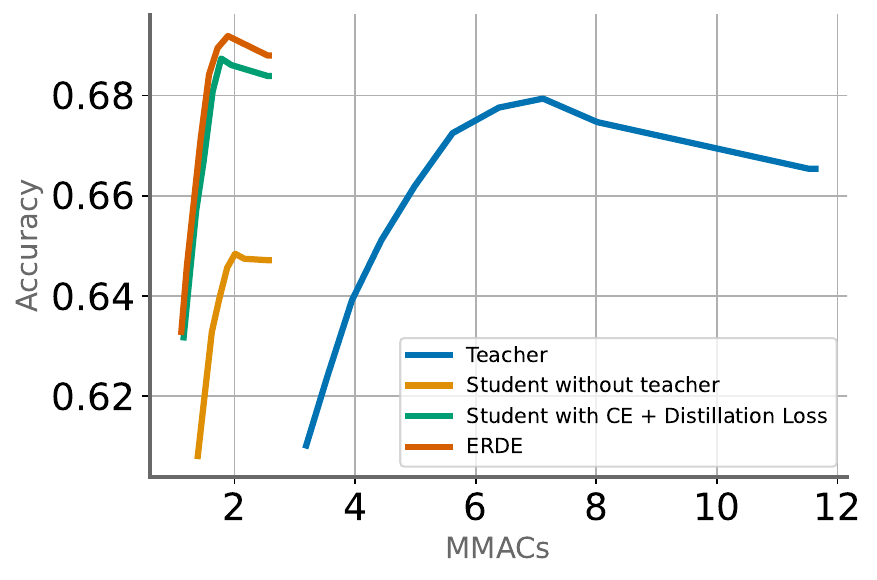}
        \caption{CIFAR100}
        \label{fig:cifar100}
    \end{subfigure}
    \begin{subfigure}{\linewidth}
        \includegraphics[width=1\textwidth]{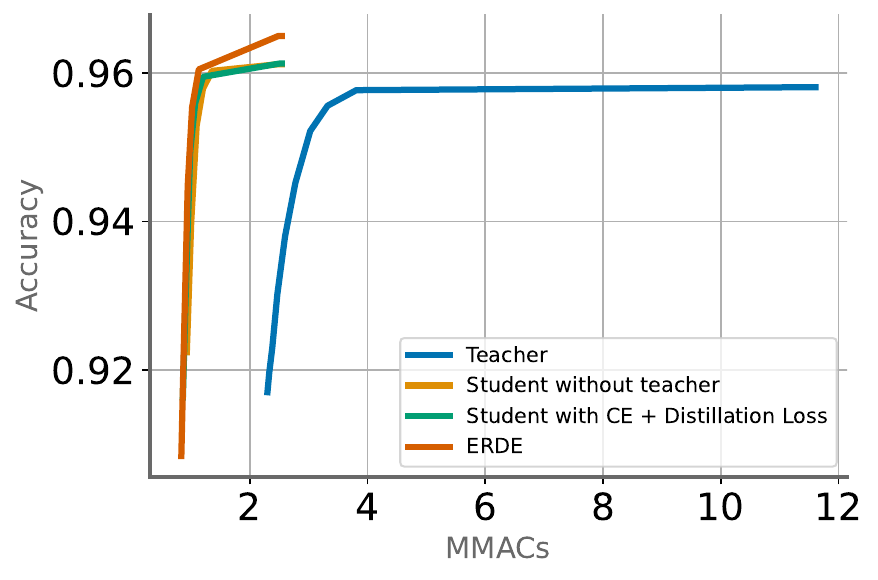}
        \caption{SVHN}
        \label{fig:svhn}
    \end{subfigure}\hfill
    \centering
    \caption{Test accuracy as a function of average MACs for different datasets and training strategies using ResNet34 as teacher and ResNet10 as student.}
    \vspace{-1em}
\end{figure}

However, sometimes this is not the case as the teachers may also tend to overfit.
For CIFAR10 and SVHN, the application of knowledge distillation has been demonstrated to yield superior outcomes in comparison to those attained without this approach, as well as those achieved by the teacher. 
These results are frequently observed in other studies; however, it is noteworthy that the incorporation of our specific entropy loss during the training process has been shown to yield more favorable results in comparison to the conventional distillation loss regardless of the EE threshold and the tested teacher-student architectures.
In some cases, our approach increases the accuracy by more the 1\% point.

As illustrated in \autoref{fig:cifar10}, \ref{fig:cifar100}, and \ref{fig:svhn}, the early exit threshold has a significant impact on the performance of the models. It is noteworthy that decreasing this threshold can lead to a substantial reduction in the number of MACs without a concomitant decline in accuracy. 

Surprisingly, for CIFAR100, it appears that the best accuracy is not obtained when the early exit threshold is maximized. This could be due to some overfitting occurring at the last exit of the model.

As shown in \autoref{tab:cifar10}, \ref{tab:cifar100}, and \ref{tab:svhn}, compared to the original teacher models, the models trained with our ERDE approach have only around 10\%-40\% of MAC operations depending on the chosen architectures with very little decrease in accuracy. 
In terms of latency, we can achieve a reduction of up to 4 times.
In some cases, we may possibly go beyond this by choosing even smaller teacher models. 

%% file: 7_conclusion.tex
\section{Conclusion}

We have proposed an original method to apply knowledge distillation to an early-exit architecture. This is achieved by using a specific loss and training process. 
This approach effectively combines knowledge distillation with early exit architectures and thus leads to compact dynamic networks that can control the accuracy-complexity trade-off at run time and leverage both compression methods to reduce the complexity even further.
The efficacy of our method has been demonstrated on three distinct datasets on different CNN architectures and different teacher-student combinations. 
Our method systematically outperforms conventional knowledge distillation and is able to reduce the computational complexity up to around 10 times with negligible decrease in accuracy.

%% file: 8_acknowledgements.tex
\section{Acknowledgements}
\label{sec:acknowledgements}

This work was granted access to the HPC resources of IDRIS under the allocation 2024-25 AD011015765 and AD011014045R2  made by GENCI. 

The authors acknowledge the ANR – FRANCE (French National Research Agency) for its financial support of the RADYAL project n°23-IAS3-0002.

%% file: 9_appendix.tex
\onecolumn
\appendix
\section{Appendix}
\begin{table}[h]
    \large
    \centering
    \begin{tabular}{lllllll}
    \toprule
    Approach &Teacher &Student        &Accuracy       &\makecell{MACs \\ (M)}       &\makecell{rel. MACs \\ (M)}     &\makecell{Latency \\ (s)}\\
    \midrule[0.2pt]
teacher &ResNet18       &       &0.9153 &556.4  &100.0\%        &7.70\\
student w/o KD  &       &ResNet10       &0.9114 &254.2  &45.7\% &6.01\\
student w KD    &ResNet18       &ResNet10       &0.9275 &254.2  &45.7\% &6.05\\
ERDE ($\theta=0$)       &ResNet18       &ResNet10       &\textbf{0.9300}   &254.2  &45.7\% &5.86\\
ERDE ($\theta=0.4$)     &ResNet18       &ResNet10       &0.9082 &\textbf{116.8}  &\textbf{21.0\%} &\textbf{3.55}\\
\cdashlinelr{1-7}
teacher     &ResNet34       &       &0.8978 & 1160 & 100.0\%        &11.52\\
student w/o KD  &       &ResNet10       &0.9084 &254.2  &21.9\% &5.99\\
student w KD    &ResNet34       &ResNet10       &0.9186 &254.2  &21.9\% &6.42\\
ERDE ($\theta=0$)       &ResNet34       &ResNet10       &\textbf{0.9221} &254.2  &21.9\% &5.74\\
ERDE ($\theta=0.6$)     &ResNet34       &ResNet10       &0.8859 &\textbf{106.7}  &\textbf{9.2\%}  &\textbf{3.55}\\
\cdashlinelr{1-7}
teacher &ResNet34       &       &0.8978 &1160.8 &100.0\%        &11.5185\\
student w/o KD  &       &ResNet8        &0.9009 &195.4  &16.8\% &5.43\\
student w KD    &ResNet34       &ResNet8        &0.9057 &195.4  &16.8\% &5.99\\
ERDE ($\theta=0$)       &ResNet34       &ResNet8        &\textbf{0.9060}  &195.4  &16.8\% &5.42\\
ERDE ($\theta=0.4$)     &ResNet34       &ResNet8        &0.8856 &\textbf{103.5}  &\textbf{8.9\%}  &\textbf{3.92}\\
\cdashlinelr{1-7}
teacher &CNXT-b  &       &0.9070  &313.6  &100.0\%        &15.98\\
student w/o KD  &       &CNXT-t  &0.8859 &91.1   &29.0\% &10.57\\
student w KD    &CNXT-b  &CNXT-t  &0.8885 &91.1   &29.0\% &10.96\\
ERDE ($\theta=0$)       &CNXT-b  &CNXT-t  &\textbf{0.9007} &91.1   &29.0\% &10.62\\
ERDE ($\theta=0.6$)     &CNXT-b  &CNXT-t  &0.8936 &\textbf{37.9}   &\textbf{12.1\%} &\textbf{5.99}\\
\cdashlinelr{1-7}
teacher &CNXT-b  &       &0.907  &313.6  &100.0\%        &15.978\\
student w/o KD  &       &ResNet8        &0.9009 &195.4  &62.3\% &5.08\\
student w KD    &CNXT-b  &ResNet8        &0.9041 &195.4  &62.3\% &4.45\\
ERDE ($\theta=0$)       &CNXT-b  &ResNet8        &\textbf{0.9066} &195.4  &62.3\% &5.08\\
ERDE ($\theta=0.6$)     &CNXT-b  &ResNet8        &0.8883 &\textbf{109.4}  &\textbf{34.9\%} &\textbf{3.60}\\
    \bottomrule
    \end{tabular}
    \vspace{.7em}
    \caption{Performance comparison for CIFAR10 for different teacher-student combinations.
    Reported MMACs and latencies are averages over the whole test dataset.}
    \label{tab:cifar10}
\end{table}

\begin{table}[h]
    \centering
    \large
    \begin{tabular}{lllllll}
    \toprule
    Approach &Teacher &Student        &Accuracy       &\makecell{MACs \\ (M)}       &\makecell{rel. MACs \\ (M)}     &\makecell{Latency \\ (s)}\\
    \midrule[0.2pt]
teacher &ResNet18       &       &0.6758 &559    &100.0\%        &8.68\\
student w/o KD  &       &ResNet10       &0.6521 &256.8  &45.9\% &7.14\\
student w KD    &ResNet18       &ResNet10       &0.6702 &256.8  &45.9\% &7.29\\
ERDE ($\theta=0$)       &ResNet18       &ResNet10       &\textbf{0.6741} &256.8  &45.9\% &7.45\\
ERDE ($\theta=0.4$)     &ResNet18       &ResNet10       &\textbf{0.6726} &\textbf{172.6}  &\textbf{30.9\%} &\textbf{5.59}\\
\cdashlinelr{1-7}
teacher &ResNet34       &       &0.6654 &1163 &100.0\%        &12.91\\
student w/o KD  &       &ResNet10       &0.6471 &256.8  &22.1\% &6.95\\
student w KD    &ResNet34       &ResNet10       &0.6839 &256.8  &22.1\% &7.01\\
ERDE ($\theta=0$)       &ResNet34       &ResNet10       &\textbf{0.6880} &256.8  &22.1\% &6.96\\
ERDE ($\theta=0.6$)     &ResNet34       &ResNet10       &\textbf{0.6842} &\textbf{157.8}  &\textbf{13.6\%} &\textbf{5.21}\\
\cdashlinelr{1-7}
teacher &ResNet34       &       &0.6654 &1163.4 &100.0\%        &12.911\\
student w/o KD  &       &ResNet8        &0.6462 &198    &17.0\% &5.50\\
student w KD    &ResNet34       &ResNet8        &0.6504 &198    &17.0\% &5.51\\
ERDE ($\theta=0$)       &ResNet34       &ResNet8        &\textbf{0.6578} &198    &17.0\% &5.51\\
ERDE ($\theta=0.6$)     &ResNet34       &ResNet8        &\textbf{0.6536} &\textbf{133}    &\textbf{11.4\%} &\textbf{4.58}\\
\cdashlinelr{1-7}
teacher &CNXT-b  &       &0.7029 &314    &100.0\%        &18.71\\
student w/o KD  &       &CNXT-t  &0.6852 &91.4   &29.1\% &13.14\\
student w KD    &CNXT-b  &CNXT-t  &0.6926 &91.4   &29.1\% &13.89\\
ERDE ($\theta=0$)       &CNXT-b  &CNXT-t  &\textbf{0.7002} &91.4   &29.1\% &12.79\\
ERDE ($\theta=1.4$)    &CNXT-b  &CNXT-t  &0.6917 &\textbf{55}     &\textbf{17.5\%} &\textbf{9.10}\\
\cdashlinelr{1-7}
teacher &CNXT-b  &       &0.7029 &314    &100.0\%        &18.7115\\
student w/o KD  &       &ResNet8        &0.6462 &198    &63.1\% &4.55\\
student w KD    &CNXT-b  &ResNet8        &0.6427 &198    &63.1\% &4.70\\
ERDE ($\theta=0$)       &CNXT-b  &ResNet8        &\textbf{0.6530} &198    &63.1\% &4.79\\
ERDE ($\theta=1.4$)    &CNXT-b  &ResNet8        &0.6374 &\textbf{133}    &\textbf{42.4\%} &\textbf{3.87}\\

    \bottomrule
    \end{tabular}
    \vspace{.7em}
    \caption{Performance comparison for CIFAR100 for different teacher-student combinations.
    Reported MMACs and latencies are averages over the whole test dataset.}
    \label{tab:cifar100}
\end{table}

\begin{table}[h]
    \large
    \centering
    \begin{tabular}{lllllll}
    \toprule
    Approach &Teacher &Student        &Accuracy       &\makecell{MACs \\ (M)}       &\makecell{rel. MACs \\ (M)}     &\makecell{Latency \\ (s)}\\
    \midrule[0.2pt]
    teacher &ResNet18       &       &0.9623 &556.4  &100.0\%        &7.29\\
student w/o KD  &       &ResNet10       &0.9592 &254.2  &45.7\% &5.86\\
student w KD    &ResNet18       &ResNet10       &0.9637 &254.2  &45.7\% &6.16\\
ERDE ($\theta=0$)       &ResNet18       &ResNet10       &\textbf{0.9657} &254.2  &45.7\% &5.91\\
ERDE ($\theta=0.4$)     &ResNet18       &ResNet10       &0.9624 &\textbf{117.3}       &\textbf{21.1\%} &\textbf{3.69}\\
\cdashlinelr{1-7}
teacher &ResNet34       &       &0.9584 &1160 &100.0\%        &10.02\\
student w/o KD  &       &ResNet10       &0.9594 &254.2  &21.9\% &5.59\\
student w KD    &ResNet34       &ResNet10       &0.9606 &254.2  &21.9\% &6.35\\
ERDE ($\theta=0$)       &ResNet34       &ResNet10       &\textbf{0.9647} &254.2  &21.9\% &5.69\\
ERDE ($\theta=0.2$)     &ResNet34       &ResNet10       &0.9606 &\textbf{113.3}       &\textbf{9.8\%}  &\textbf{3.57}\\
\cdashlinelr{1-7}
teacher &ResNet34       &       &0.9584 &1160.8 &100.0\%        &10.015\\
student w/o KD  &       &ResNet8        &0.9572 &195.4  &16.8\% &4.87\\
student w KD    &ResNet34       &ResNet8        &0.9599 &195.4  &16.8\% &4.86\\
ERDE ($\theta=0$)       &ResNet34       &ResNet8        &\textbf{0.9607} &195.4  &16.8\% &4.87\\
ERDE ($\theta=0.2$)     &ResNet34       &ResNet8        &0.9566 &\textbf{105.3}       &\textbf{9.1\%}  &\textbf{3.45}\\
\cdashlinelr{1-7}
teacher &CNXT-b  &       &0.9476 &313.6  &100.0\%        &15.96\\
student w/o KD  &       &CNXT-t  &0.9468 &91.1   &29.0\% &10.71\\
student w KD    &CNXT-b  &CNXT-t  &0.9545 &91.1   &29.0\% &10.98\\
ERDE ($\theta=0$)       &CNXT-b  &CNXT-t  &\textbf{0.9568} &91.1   &29.0\% &10.69\\
ERDE ($\theta=1.4$)    &CNXT-b  &CNXT-t  &\textbf{0.9556} &\textbf{23.0}        &\textbf{7.5\%}  &\textbf{4.86}\\
\cdashlinelr{1-7}
teacher &CNXT-b  &       &0.9476 &313.6  &100.0\%        &15.957\\
student w/o KD  &       &ResNet8        &0.9572 &195.4  &62.3\% &4.867\\
student w KD    &CNXT-b  &ResNet8        &0.9549 &195.4  &62.3\% &5.010\\
ERDE ($\theta=0$)       &CNXT-b  &ResNet8        &\textbf{0.9574} &195.4  &62.3\% &5.089\\
ERDE ($\theta=1.6$)    &CNXT-b  &ResNet8        &\textbf{0.9557} &\textbf{110.1}       &\textbf{35.1\%} &\textbf{3.963}\\

    \bottomrule
    \end{tabular}
    \vspace{.7em}
    \caption{Performance comparison for SVHN for different teacher-student combinations.
    Reported MMACs and latencies are averages over the whole test dataset.}
    \label{tab:svhn}
\end{table}